\documentclass{article}
\usepackage{iclr2024_conference,times}

\usepackage[
    pdftex,
    colorlinks,
    linkcolor=black,
    citecolor=purple,
    urlcolor=black,
    filecolor=black,
    pagebackref    
]{hyperref}
\usepackage{url}
\usepackage{subcaption}
\usepackage{booktabs}
\usepackage{float}
\usepackage{multirow}
\usepackage{graphicx}
\usepackage{mathtools}
\usepackage{amsthm}
\usepackage{bm}

\usepackage{fancyhdr}
\fancypagestyle{style2}{
\fancyhf{}
\fancyhead[L]{Preprint, submission in preparation}
\fancyhead[R]{page \thepage}
}

\title{
HOSC: A Periodic Activation Function for Preserving Sharp Features in Implicit Neural Representations
}

\author{%
Danzel Serrano\\
\small New Jersey Institute of Technology\thanks{\texttt{https://www.njit.edu/}}\\
\small \texttt{ds867@njit.edu}
\AND 
Jakub Szymkowiak\\
\small Adam Mickiewicz University\thanks{\texttt{https://www.amu.edu.pl/}}\\
\small IDEAS NCBR\thanks{\texttt{https://www.ideas-ncbr.pl/}}\\
\small \texttt{jakub.szymkowiak@ideas-ncbr.pl}
\AND 
Przemyslaw Musialski\\
\small New Jersey Institute of Technology\\
\small IDEAS NCBR\\
\small \texttt{przem@njit.edu}
}

\DeclarePairedDelimiter{\br}{(}{)}

\DeclarePairedDelimiter{\arr}{[}{]}
\DeclarePairedDelimiter{\norm}{\|}{\|}

\DeclareMathOperator{\sdf}{sdf}
\DeclareMathOperator{\HOSC}{HOSC}
\DeclareMathOperator{\sign}{sign}
\DeclareMathOperator{\relu}{ReLU}

\renewcommand{\sharp}{\beta}

\newcommand{\N}{\mathrm{\mathbf{N}}}
\newcommand{\R}{\mathrm{\mathbf{R}}}
\newcommand{\E}{\mathrm{\mathbf{E}}}

\newcommand{\Image}{\mathrm{\mathbf{I}}}

\newcommand{\bff}{\mathbf{f}}

\newcommand{\bfs}{\mathbf{s}}

\newcommand{\bfx}{\mathbf{x}}

\newcommand{\bX}{\mathbf{X}}

\newcommand{\btheta}{\boldsymbol{\theta}}

\iclrfinalcopy %
\begin{document}
\pagestyle{style2}

\maketitle

\begin{abstract}
    Recently proposed methods for implicitly representing signals such as images, scenes, or geometries using coordinate-based neural network architectures often do not leverage the choice of activation functions, or do so only to a limited extent. 
    In this paper, we introduce the Hyperbolic Oscillation function (HOSC), a novel activation function with a controllable sharpness parameter. 
    Unlike any previous activations, HOSC has been specifically designed to better capture sudden changes in the input signal, and hence sharp or acute features of the underlying data, as well as smooth low-frequency transitions. 
    Due to its simplicity and modularity, HOSC offers a plug-and-play functionality that can be easily incorporated into any existing method employing a neural network as a way of implicitly representing a signal. 
    We benchmark HOSC against other popular activations in an array of general tasks, empirically showing an improvement in the quality of obtained representations, provide the mathematical motivation behind the efficacy of HOSC, and discuss its limitations.
    \end{abstract}
    
    \section{Introduction}
    An increasingly common scenario in learning visual data representations is approximating a structured signal $\bfs \colon \R^k \to \R^m$ via a coordinate-based neural network $\bff_{\btheta}$ parametrized by a set of parameters $\btheta \in \R^p$. 
    These representations, known as implicit neural representations (INRs), are fully differentiable and offer numerous advantages over traditional counterparts such as meshes or pixel grids in optimization tasks, often requiring significantly less memory.
    
    INRs are versatile in their application, capable of representing a variety of types of objects, including audio signals ($k,m=1$), images ($k=2$, $m=1$ or $m=3$), radiance fields ($k=5$, $m=4$), geometries ($k=2$ or $k=3$, $m=1$), and parametrzied curves ($k=1$, $m>1$). 
    For instance, to represent the geometry of a 3D object, one would obtain a dataset of evaluations $\bX = \{ (\bfx, \bfs(\bfx)) \}$ of the signed distance function $\bfs(\bfx) = \sdf(\bfx)$ with respect to the surface of that object, and find the values of parameters $\btheta$ that minimize the reconstruction loss:
    \[
        \mathcal{L}(\btheta) = \min_{\btheta} \ \E_{(\bfx, \bfs(\bfx)) \sim \bX} \arr*{ \norm*{\bff_{\btheta}(\bfx) - \bfs(\bfx)}_2 + \Psi(\btheta ) } \, ,
    \]
    where $\Psi(\btheta)$ denotes a regularizer. 
    Instead of a regression, this task could also be posed as a classification problem, where the signal $\bfs(\bfx)$ takes values in the discrete set $\{0,1\}$, representing an occupancy field. 
    In general, defining an appropriate domain, codomain, loss function, and regularization is a problem-specific research challenge.
    
    \begin{figure}[t]
        \centering
        \begin{subfigure}{0.245\linewidth}
             \centering
             \includegraphics[width=\linewidth]{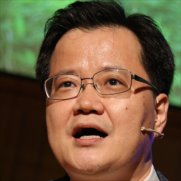}\\
             \caption{Ground Truth}
         \end{subfigure}
         \hfill     
         \begin{subfigure}{0.245\linewidth}
             \centering
             \includegraphics[width=\linewidth]{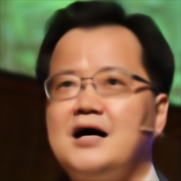}\\
             \caption{$\relu(x)$}
         \end{subfigure}
         \hfill
         \begin{subfigure}{0.245\linewidth}
             \centering
             \includegraphics[width=\linewidth]{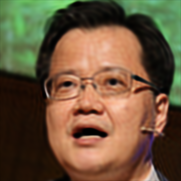}\\
             \caption{$\sin(x)$}
         \end{subfigure}
         \hfill
         \begin{subfigure}{0.245\linewidth}
             \centering
             \includegraphics[width=\linewidth]{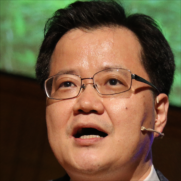}\\
             \caption{$\HOSC(x; 8)$}
         \end{subfigure}
        \caption{
        Reconstruction of an image using an MP running different activation functions.
        The process involved training a five-layer coordinate-based MLP with a width of $256$ for $100$ iterations for each of the activations.
        No positional encoding and no frequency initialization has been used.
        }
        \label{fig:janson}
    \end{figure}

    Importantly, INRs introduce a new paradigm in training neural networks. 
    In classical applications of neural networks, such as prediction, the goal is to approximate a function $\bff$ given its noisy evaluations $\bff(\bfx)$ at sparsely sampled datapoints $\bfx$. 
    One of the challenges is thus not to overfit the approximation to the noise present in the training data.
    On the contrary, for INRs, we assume the data is noise-free and more regularly sampled, and aim to encode this into the network's parameters, implying that in this context overfitting is actually desirable for capturing high-frequency details of the signal.
    
    However, popular activation functions such as ReLU are biased towards capturing lower frequencies, which is beneficial in prediction tasks, but hinders their capability to accurately represent sharp features of signals when applied as INRs.
    Three primary strategies to approach this problem while remaining in the INRs framework have been developed:
    \begin{itemize}
        \item \textbf{Hybrid representations.} 
        Methods like ACORN \citep{Martel2021Acorn}, InstantNGP \citep{Mller2022InstantNG} and TensoRF \citep{Chen2022TensoRFTR} use neural networks to achieve highly detailed representations of complex signals, such as gigapixel images and radiance fields.
        However, they also rely on traditional data structures, and hence require storing some sort of raw data.
        This notably enlarges their memory footprint compared to just storing the parameters of an MLP, and results in them not being fully differentiable.
        \item \textbf{Positional encoding.} 
        Fourier Feature Networks (FFNs) \citep{Tancik2020FourierFL} employ positional encoding, which has been shown to accelerate the learning of higher-frequency features.
        Such encodings, if sampled densely, become extremely memory inefficient, and therefore require sampling a predefined distribution.
        This introduces more stochasticity to the model, as well as the need to tune the distribution's parameters manually.
        \item \textbf{Periodic activations.}
        Sinusoidal Representation Networks (SIRENs) proposed by \citet{Sitzmann2020ImplicitNR} are multi-layer perceptrons (MLPs) that utilize $\sin(x)$ instead of ReLU as their activation function.
        Consequently, they remain fully differentiable and offer a compact representation of the signal.
        While SIRENs demonstrated a significant improvement over ReLU, they struggle to capture high-frequency details in problems like shape representations, and are not-well suited for methods such as \citep{Mildenhall2020NeRF}.
    \end{itemize}

    In this paper, we introduce a new periodic parametric activation function --- the \textbf{Hyperbolic Oscillation} activation function (\textbf{HOSC}), defined as $\HOSC(x; \sharp) = \tanh (\sharp \sin x)$.
    Here, $\sharp > 0$ is a controllable sharpness parameter, enabling HOSC to seamlessly transition between a smooth sine-like wave and a square signal.
    Similarly to SIREN, an MLP running the HOSC activation function is fully differentiable and inexpensive memorywise. 
    However, the HOSC's sharpness parameter $\sharp$ allows it to much more accurately capture sudden or sharp jumps, and hence preserve high-frequency details of the signal.
    Moreover, since HOSC is differentiable with respect to $\sharp$, the sharpness can be adjusted automatically alongside the reset of the parameters, a method to which we refer as \textbf{Adaptive HOSC} or \textbf{AdaHOSC}.
    
    Our extensive empirical studies show that HOSC consistently outperforms ReLU and SIREN across an array of benchmarking tasks.  
    These tasks encompass fitting random signals, images of random square patches, photos, gigapixel images, and 2D \& 3D SDF.
    In summary, HOSC provides an easy-to-implement method allowing simple MLPs to achieve high level of detail in signal encoding tasks without loosing differentiability or increasing memory footprint, and it does this without the need for positional encoding.
    
    \section{Related work}
    \subsection{Implicit Neural Representations}
    Currently, INRs are gaining a lot of attention in visual computing research \citep{Xie2021NeuralFI}.
    Their applications are widespread, and encompass image processing \citep{Tancik2020FourierFL}, radiance fields \citep{Mildenhall2020NeRF}, 3D shape modeling \citep{Park2019DeepSDFLC}, audio and video compression \citep{Lanzendrfer2023SiameseSA, Chen2021NeRVNR}, physics-informed problems \citep{Raissi2019PhysicsinformedNN}, and solving PDEs \citep{Sitzmann2020ImplicitNR, Li2020FourierNO}. 
    There are many reasons for choosing INRs over classical data structures:
    \begin{itemize}
        \item \textbf{Differentiability.} Given their differentiable nature, INRs offer an immediate advantage over classical, non-differentiable methods in optimization and deep learning tasks.
        \item \textbf{Compactness.} INRs often require less memory, as storing the parameters and hyperparameters of a neural network is typically less memory-intensive than storing raw data.
        \item \textbf{Continuous representation.} In principle, due to their generalization capability, neural networks enable the representation of data with arbitrary precision, making resolution a less significant issue \citep{Chen2020LearningCI}.
    \end{itemize}
    For a more comprehensive review of the INR literature, we refer to the recent surveys by \citet{Tewari2020StateOT}, \citet{Tewari2021AdvancesIN}, and \citet{Xie2021NeuralFI}.
    
    \begin{figure}[t]
        \centering
        \begin{subfigure}{0.325\linewidth}
             \centering
             \includegraphics[width=\linewidth]{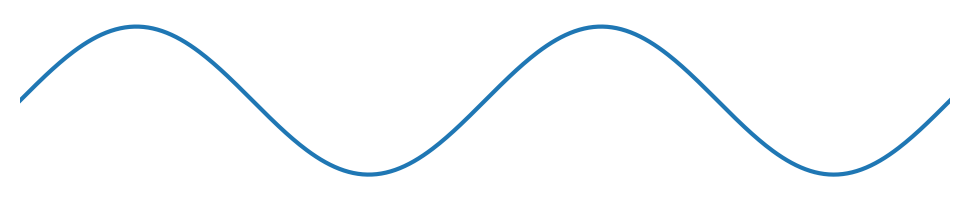}\\
             \caption{$\sin x$}
         \end{subfigure}
         \begin{subfigure}{0.325\linewidth}
             \centering
             \includegraphics[width=\linewidth]{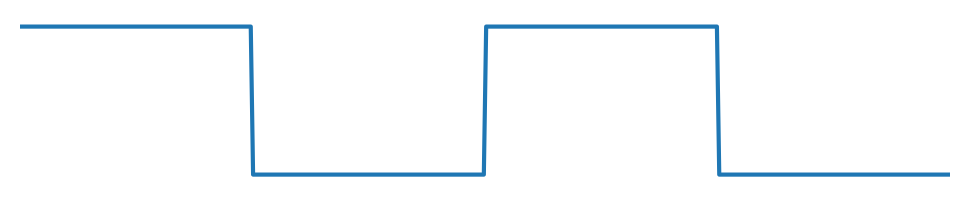}
             \caption{$\sign(\sin x)$}
         \end{subfigure}
         \begin{subfigure}{0.325\linewidth}
             \centering
             \includegraphics[width=\linewidth]{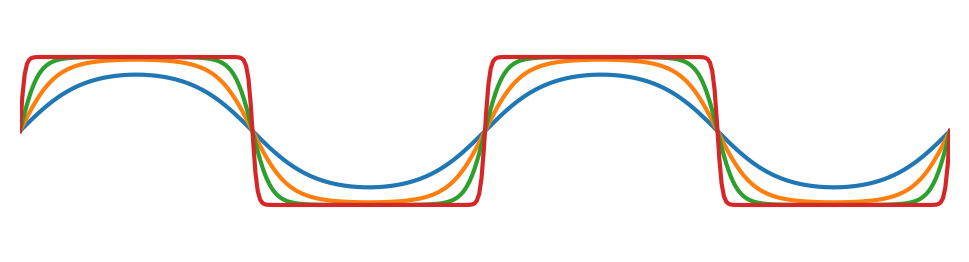}\\
             \caption{$\HOSC(x; \sharp)$}
         \end{subfigure}
        \caption{
        Comparison of the sine, square, and HOSC waves for different values of the sharpness parameter $\sharp \in \{1, 2, 4, 16\}$.
        As $\sharp$ increases, HOSC starts to resemble a square wave.
        }
        \label{fig:hosc_wave}
    \end{figure}

    \textbf{Shape and geometry representation.}
    Classical methods of shape and geometry representation include voxel grids, polygonal meshes and point clouds.
    However, all of these methods suffer from limitations.
    Voxel grids are subject to the curse of dimensionality, which makes them inefficient in handling high-resolution data.
    Moreover, manipulating voxel grids and dense meshes can be computationally intensive \citep{Xiao2020ASO, Kato2017Neural3M}.
    Meshes are also prone to errors, and designing a mesh can be quite time-consuming for human creators.
    As for point clouds, they do not encode topological information \citep{Kato2017Neural3M}.
    These issues have prompted the exploration of INRs in the context of shape and geometry modeling.
    The seminal work by \citet{Park2019DeepSDFLC} has demonstrated that INRs are capable of accurately representing surfaces as signed distance functions.
    Further research in this direction has been conducted by \citet{Atzmon2019SALSA}, \citet{Michalkiewicz2019ImplicitSR} and \citet{Gropp2020ImplicitGR}.
    Another option is presented by occupancy networks, which model the shape as the decision boundry of a binary classifier implemented as a neural network \citep{Mescheder2018OccupancyNL, Chen2018LearningIF}.
    
    \textbf{Encoding appearence.}
    In addition to encoding geometry, coordinate-based neural networks are also capable of representing the appearence aspects.
    For instance, Texture Fields \citep{Oechsle2019TextureFL} enable coloring any 3D shape based on an image.
    Methods such as LIIF \citep{Chen2020LearningCI} and ACORN \citep{Martel2021Acorn} are effective in representing high-resolution gigapixel images. 
    Furthermore, by addressing the inverse problem, Neural Radiance Fields \citep{Mildenhall2020NeRF} allow for reconstruction of multidimensional scenes from a collection of 2D images.
    Other significant contribution in this area include \citep{Mller2022InstantNG, Chen2023DictionaryFL, Chen2022TensoRFTR, Martel2021Acorn}.
    A lot of these and similar methods are hybrid representations that combine neural networks with classical non-differentiable data structres.
    As such, they are not directly related to HOSC, which primarily focuses on fully differentiable architectures.
    
    \subsection{Activation Functions and Periodicity}
    \textbf{Activation functions.}
    Activations are essential for neural networks to be able to model non-linear relationships.
    Early activation functions include the Logistic Sigmoid, Hyperbolic Tangent, and Rectified Linear Unit (ReLU). 
    Thanks to their low-frequency bias, they are able to deal with the noise present in the training data, possess generalization capabilities, and thus excel in applications such as prediction.
    In contrast to these early non-linearities, more recently proposed activation functions such as SWISH \citep{Ramachandran2018SearchingFA}, PReLU \citep{He2015DelvingDI}, SReLU \citep{Jin2015DeepLW}, and MPELU \citep{Li2016ImprovingDN} incorporate one or more parameters, which are optimized during training along with the rest of the network's parameters.
    For a more in-depth survey on activation functions, refer to \citep{Dubey2021ActivationFI, Apicella2020ASO, Karlik2011PerformanceAO}.
    
    \textbf{Periodicity in neural networks.}
    All the non-linearities mentioned in the previous section are non-periodic.
    Altough less common, periodic activations have been studied for many years.
    Early work by \citet{Sopena1999NeuralNW} and \citet{Wong2002HandwrittenDR} analyzed their performance in classification problems.
    A more recent study by \citet{Parascandolo2017TamingTW} investigated which tasks are particularly well-suited to periodic activations and where they may face challenges.
    In \citep{Lapedes1987NonlinearSP}, the authors use an MLP with sine activation for signal modeling, drawing a direct connection to the Fourier transform.
    A significant contribution in the field of INRs is the SIREN architecture \citep{Sitzmann2020ImplicitNR}, which employes sine activation to solve PDEs and encode images and videos. 
    Various aspects of periodic activations have also been studied by \citet{Ramasinghe2021BeyondPT}.
    An alternative approach to introducing periodicity has been explored by \citet{Tancik2020FourierFL}, who generalize positional encoding to coordinate-based MLPs.

    \begin{figure}[t]
        \centering
        \includegraphics[width=\textwidth]{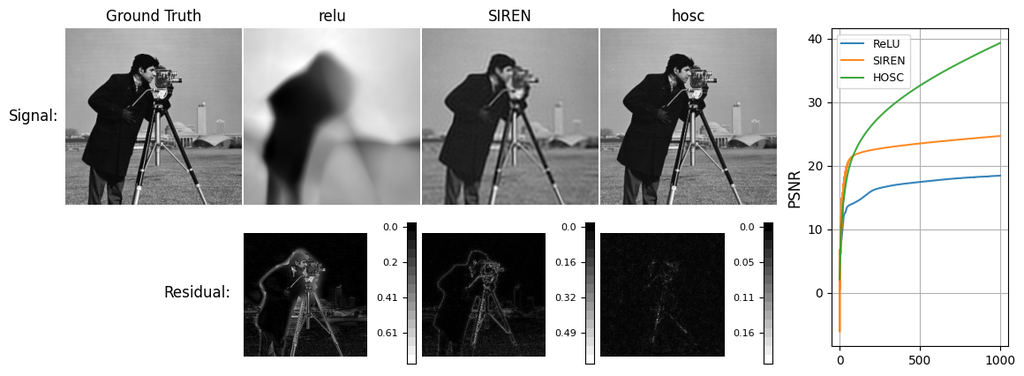}\\
        \includegraphics[width=\textwidth]{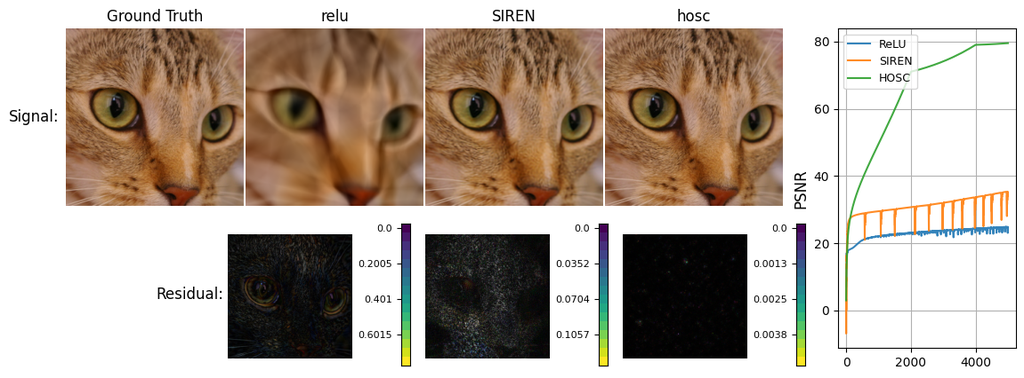}
        \caption{Comparison of a ReLU, SIREN, and HOSC fitting the 'Cameraman' image for 1000 epochs and a high-frequency detail 'Cat' image for 5000 epochs. The plot to the right shows PSNRs of the model to ground truth per epoch of training; for the 'Cat' we used adaptively scheduled learning rate. Below each resulting model is the residual difference from the ground truth signal.}
        \label{fig:cameraman}
    \end{figure}

    \section{Our Contribution}
    \subsection{HOSC}
    In this paper, we propose a novel periodic parametric activation function designed specifically for fitting INRs --- the Hyperbolic Oscillation activation function, or HOSC.
    It is defined as
    \begin{equation*}
        \HOSC(x; \sharp) = \tanh(\sharp \sin x) \, ,
    \end{equation*}
    where $\sharp > 0$ is the sharpness parameter, controlling the extent to which the resulting wave resembles a square wave.
    This phenomenon is illustrated in Figure \ref{fig:hosc_wave}. 
    In fact, given that $\lim_{\sharp \to \infty} \tanh(\sharp x) = \sign(x)$ for all $x \in \R$, we know that $\lim_{\sharp \to \infty} \HOSC(x; \sharp) = \sign(\sin x)$, so indeed HOSC approaches the square wave pointwise in the infinite sharpness limit.
    The rapid amplitude changes around $x = n \pi$ for $n \in \N$ at high values of $\sharp$ enable HOSC to model acute features of the signal.
    Conversely, smooth transitions at lower $\sharp$ values allow it to capture low-frequency components instead.
    
    \subsection{AdaHOSC}
    Importantly, HOSC is differentiable not only with respect to the input $x$, but also with respect to the sharpness parameter $\sharp$:
    \[
        \partial_{\sharp} \HOSC(x; \sharp) = \sin(x) \br*{ 1 - \HOSC^2(x; \sharp) } \, .
    \]
    This property allows an MLP using HOSC to optimize the sharpness parameter during training, rather than fixing it as a hyperparameter. 
    When the sharpness factor $\sharp$ is dynamically optimized, we refer to the resulting activation function as AdaHOSC, an acronym for Adaptive HOSC.

    \section{Experimental results}
    In this section, we experimentally assess the performance of HOSC in various benchmarking tests and compare it to ReLU and SIREN.
    More experimental results can be found in the Appendix.
    
    \subsection{Representing images ($\bfs \colon \R^2 \to \R$ or $\R^3$)}
    An image can be conceptualized as a function $\Image \colon \R^2 \to \R^n$, where $n=1$ (for black and white images) or $n=3$ (in case of the colored images), mapping pixel coordinates to their corresponding color intensities.
    To construct an INR, one commonly approximates the function $\Image$ with an MLP, training it on all the available coordinate-color value pairs $((x,y), \Image(x,y))$.

    In Figures \ref{fig:janson} and \ref{fig:cameraman} we present the results of fitting photos with an MLP running the HOSC activation.
    In Figure \ref{fig:cameraman}, the black and white cameraman image is fitted for $1000$ epochs, demonstrating that an MLP employing HOSC activation achieves a higher PSNR quicker than a ReLU-MLP or SIREN.
    We also note that we employed a linear step-wise learning rate scheduler with a rate of $\gamma = 0.1$ every $2000$ epochs, as the HOSC-MLP begins to exhibit an extreme oscillatory convergence at high PSNR values without this adjustment. 
    Although we confirm that the Gaussian activation performs better than the SIREN model in this case, our findings reveal that a HOSC-MLP achieves a significantly higher PSNR compared to both.
    
\begin{figure}[t]
        \centering
        \includegraphics[width=\linewidth]{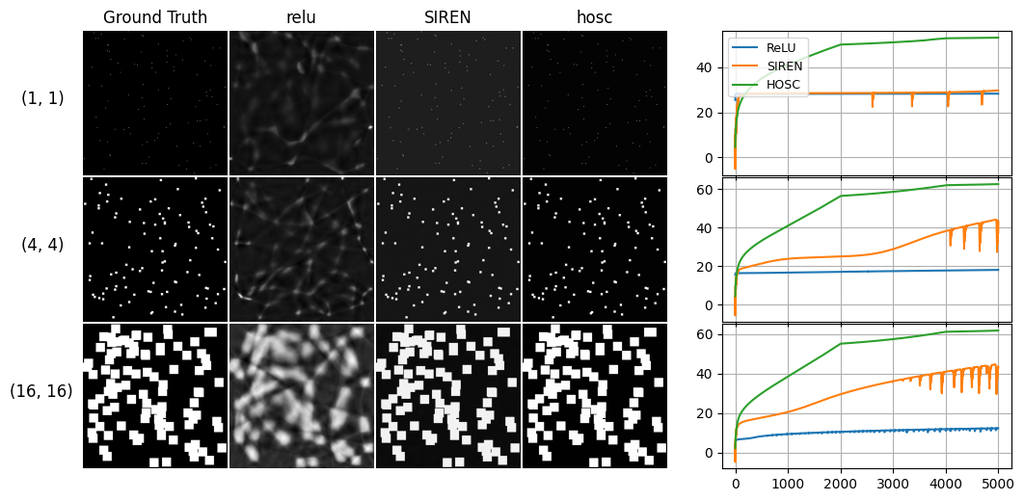}
        \caption{Comparison of a ReLU, SIREN, and HOSC fitting an image of random square patches of dimension 1x1, 4x4, and 16x16. The plot to the right shows PSNRs of the model to ground truth per epoch of training.}
        \label{fig:img_square_patches_1_4_16}
    \end{figure}

    Figure \ref{fig:img_square_patches_1_4_16} reveals more interesting results.
    This experiment evaluates the performance of the HOSC on images with varying frequency content.
    For each activation, a four-layer MLPs with a width of $256$ was trained on a $256 \times 256$ black images, each containing $100$ randomly placed white square patches, over $5000$ epochs.
    Patch sizes used in the experiment are $1 \times 1$, $4 \times 4$, and $16 \times 16$.
    For the SIREN model, we adopted the same weight initialization and a frequency factor of $30$, as detailed in \citep{Sitzmann2020ImplicitNR}.
    For the HOSC-MLP, we use a sharpness factor schedule, where sharpness varies across layers with values $\sharp_i = [2, 4, 8, 16]$.
    Additionally, a frequency factor of $30$ is used in the first layer, followed by a factor of $1.0$ in the subsequent layers.
    
    As anticipated, the ReLU network struggles to accurately capture the ground truth signal, resulting in a low converging PSNR.
    Interestingly, as the patch size increases, ReLU's peak PSNR decreases. In contrast, for SIREN, the peak PSNR increases as the ground truth signal losses sharp frequencies.
    The HOSC-MLP surpasses both of them, and is able to accurately represent the ground truth signal regardless of patch sizes. 
    Notably, the PSNR values for the HOSC-MLP, after being trained for $5000$ epochs, significantly exceed those of both the ReLU-MLP and SIREN models.
    
    \subsubsection{Gigapixel images}
    \begin{figure}[ht]
        \centering
        \includegraphics[width=\linewidth]{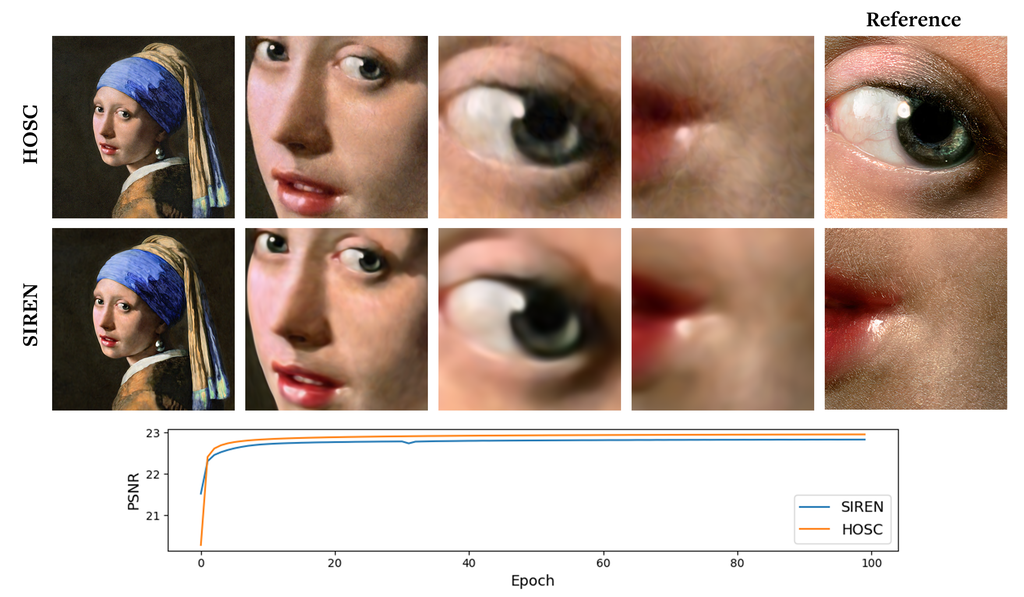}
        \caption{
        Top: Results of fitting a HOSC and a SIREN model to a high-resolution image.
        Bottom: A plot of PSNR per epoch for both methods.
        }
        \label{fig:hosc_siren_gigapixels_psnr}
    \end{figure}
    
    In this experiment, we compare the performances of SIREN and HOSC models on the task of Gigapixel Image Approximation, where the target signal is an RGB Image of extremely high-resolution. 
    In this case, the SIREN model is of depth $4$, with a width of $256$, and each sinusoidal activation has a frequency of $30$. 
    The HOSC model has the same depth and width, however, the activations have a sharpness of $\sharp=8$, while only the first activation has a frequency of $30$, and the rest has a frequency of $1$. 
    We follow the same initialization scheme as \citet{Sitzmann2020ImplicitNR} for both the SIREN and HOSC models.
    
    Results of fitting both models to an image of resolution $9302 \times 8000 \times 3$ for $100$ epochs are shown in Figure \ref{fig:hosc_siren_gigapixels_psnr}. 
    Although both models do not look perceptually close to the zoomed in reference photo, it is apparent that a model equipped with HOSC is able to retain sharper features for the high-resolution image whereas the SIREN essentially learns a smooth interpolation. 
    This fact is also reinforced by analyzing the PSNR plots, where HOSC beats SIREN even at the first few epochs.
    
    \subsection{SDFs ($\bfs \colon \R^2$ or $\R^3 \to \R$)}
    The signed distance $\sdf_{\Omega}(\bfx)$ with respect to a shape $\Omega \subset \R^n$ is defined as
    \[
        \sdf_{\Omega}(\bfx) = 
            \begin{cases}
                +\rho(\bfx, \partial \Omega) &\text{if } \bfx \not \in \Omega \, , \\
                -\rho(\bfx, \partial \Omega) &\text{if } \bfx \in \Omega \, ,
            \end{cases}
    \]
    for all the coordinates $\bfx \in \R^n$. 
    Consequently, by training a neural network $\bff_{\btheta}$ to approximate $\sdf_{\Omega}$, we can approximate the shape's boundary as the zero-level set $\{ \bfx \in \R^n \mid \bff_{\btheta}(\bfx) = 0 \}$.
    
    In this section, we consider the case where $n=2$ and $n=3$ and train MLPs on a dataset $\{ (\bfx, \sdf(\bfx)) \}$ comprised of coordinate-SDF evaluation pairs.
    Our experimental exploration aims to identify any patterns that emerge when varying the depth and width of the MLPs, as well as adjusting the sharpness factor $\sharp$ (2D SDF).
    
    In Figure \ref{fig:3d_sdf_1}  we present a comparison of AdaHOSC to ReLU and SIREN.
    For a fair comparison, we run the same MLP architecture ($5$ hidden layers with $256$ width) for $20$ epochs only changing the activation. 
    AdaHOSC uses the initial value of $\sharp=8$.
    We find that AdaHOSC provides a much higher quality of representation, as demonstrated by the IoU (intersection over union) values, suggesting faster convergence time compared to SIREN.
    
\begin{figure}[h]
        \centering
        \begin{subfigure}{0.95\linewidth}
            \includegraphics[width=\textwidth]{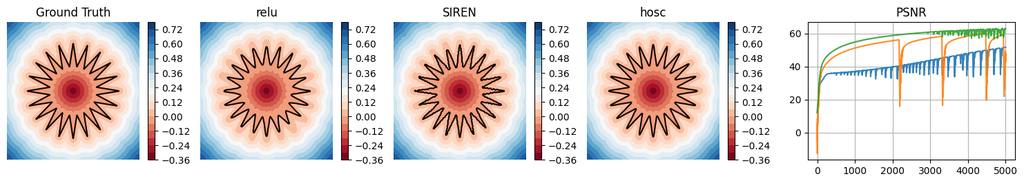}
        \end{subfigure}
        \caption{
            Comparison of coordinate-based MLPs fitting an SDF of a regular star.
            Top: SDFs learned by the models running different activations.
            Bottom: Heatmap illustrating the maximum PSNR values for HOSC-MLPs with different topologies (depth and width of layers) and sharpness factors.
        }
        \label{fig:2d_sdf_star}
    \end{figure}

    Moreover, the results of the 2D SDF experiment are illustrated in Figure \ref{fig:2d_sdf_star}.
    In this experiment, we train four-layer $512$ width MLPs on $20$ SDF evaluations of a regular star shape.
    Similar to image fitting, HOSC's performance surpasses that of ReLU and SIREN.
    Moreover, our findings indicate that deep HOSC-MLPs achieve higher PSNR values.
    Regardless of depth, it is observed that, depending on the width, greater $\sharp$ values enable HOSC to more accurately represent the shape, as evidenced by the PSNR values.
    This observation further supports the hypothesis that HOSC can effectively represent signals with high-frequency content (including discontinuities) when the sharpness factor is large.
    
    \section{Conclusions and Discussion}
     In this paper, we have introduced the Hyperbolic Oscillation activation function $\HOSC(x; \sharp) = \tanh(\sharp \sin x)$, a new periodic parametric activation that has been designed to be particularly effective in preserving sharp features in INRs.
    Additionally, in Section 4, we presented experimental results that evaluate the performance of the HOSC function in comparison to existing approaches.
    Our findings revealed that an MLP employing the HOSC activation with a suitably chosen or automatically-optimized sharpness parameter $\sharp$ consistently outperforms identical structure MLPs using ReLU and SIREN activations, and achieve the same level of accuracy in neural signal encoding problems as Fourier Feature Networks.
    HOSC thus offers a simple, fully differentiable and compact high-quality signals representation method with no need for hyperparameter tuning.
    However, we have identified scenarios where HOSC is clearly not the optimal choice, which we will explore in the following discussion.
    
    \textbf{Spectral bias.} 
    Different problems require a different spectral bias.
    While for signal encoding, where we assume that there is very little or no noise present in the data, fitting high-frequency components of the signal is advantageous.
    Conversely, for capturing only the general trends from a set of noisy data, a low-frequency bias can help avoid overfitting to noise.
    Naturally, this constraints the application of HOSC and other periodic activations in settings that require generalization beyond the observed datapoints.
    For instance, in our experiments we found that both HOSC and SIREN activations underperform when applied to Neural Radiance Fields \citep{Mildenhall2020NeRF}, which perform best with various types of positinoal encoding, either in freqnecty domain, or parametric encoding combined with spatial data strcutres (also denoted as hybrid) and ReLU activations.
    
    \textbf{Solving PDEs.}
    Cooridnate-based neural networks have been applied to solving PDEs in physics \citep{Raissi2019PhysicsinformedNN}. 
    However, we observe that HOSC is not particularly suited for these types of problems, compared to e.g. SIREN architecture. 
    We attribute this limitation to the increasing complexity found in subsequent derivatives of HOSC.
    While the derivatives of a fully-connected SIREN layer remain SIRENs \citep{Sitzmann2020ImplicitNR}, enabling it to accurately fit both the signal and its derivatives, the situation is more convoluted.
    As a result, HOSC preserves the signal's sharp features but at the expense of derivative information.
    
    \textbf{Hybrid and parametric positional encoding.}
    In our experiments we apply HOSC in signal encoding scenarios, like images, giga-pixel images, and 3D SDFs, where a HOSC-MLP achieves similar reconstruction quality, however, not the timings of highly optimized methods as InstantNGP \citep{Mller2022InstantNG}, ACORN \citep{Martel2021Acorn}, grid-based Dictionary Fields \citep{Chen2023DictionaryFL}, or TensoRF \citep{Chen2022TensoRFTR}. 
    There methods shorten the training and inference times at the cost of a higher memory footprint and a more sophisticated implementation. 
    In contrast, a simple MLP is much easier to implement and storing its parameters is far less demanding in terms of memory. 
    Finally, it is important to note that, in principle HOSC can be utilized in hybrid representations as well, whenever a coordinate-based MLP is used to overfit a signal.
    
    \textbf{Architecture design.} 
    A deeper understanding of how neural networks represent implicitly encoded signals may also provide greater insights into the design of non-MLP network architectures, enabling HOSC to fully leverage their capabilities.
    The research presented in \cite{Ramasinghe2021BeyondPT} offers intresting ideas relevant to this context.
    Furthermore, HOSC naturally fits in the Factor Fields framework \citep{Chen2023FactorFA}.
    More specifically, we can let any factor $\bfs$ in a Factor Field be modeled with a HOSC MLP, and vary the sharpness $\sharp$ across the factors.
    This means that the framework could potentially be used to develop novel representation methods using HOSC, possibly combined with other architectures.
    
    \newcommand{\mysubcaption}[3]{%
        \caption{\parbox{0.99\linewidth}{\footnotesize \centering #1 \\ #2, #3}}
    }
    \begin{figure}[H]
        \centering
        \captionsetup[subfigure]{labelformat=empty}
    
        \begin{subfigure}{0.244\linewidth}
        \centering
            \includegraphics[height=4cm]{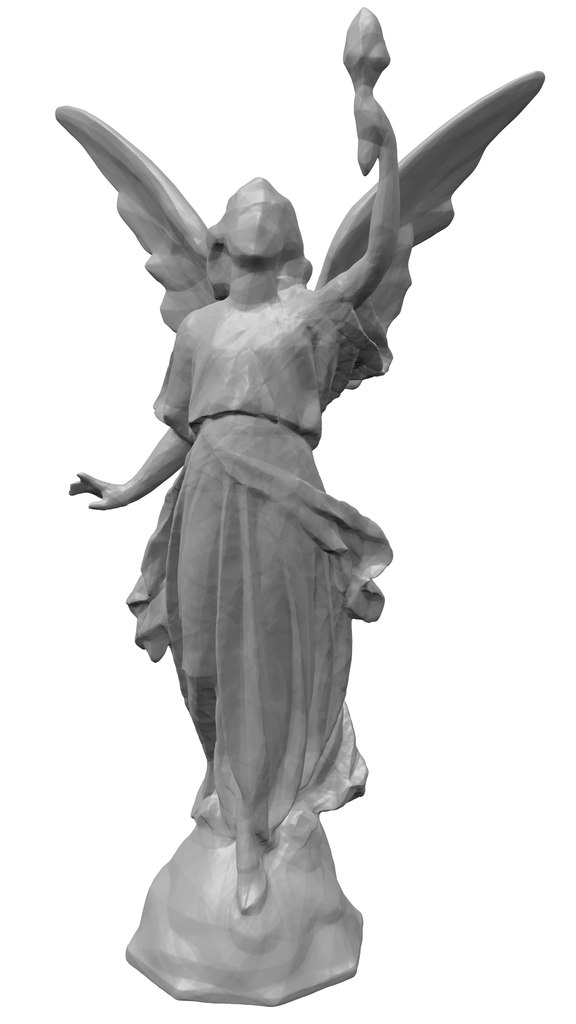}
            \mysubcaption{Lucy}{IoU}{train time}        
        \end{subfigure}\hfill
        \begin{subfigure}{0.244\linewidth}
            \includegraphics[width=\linewidth]{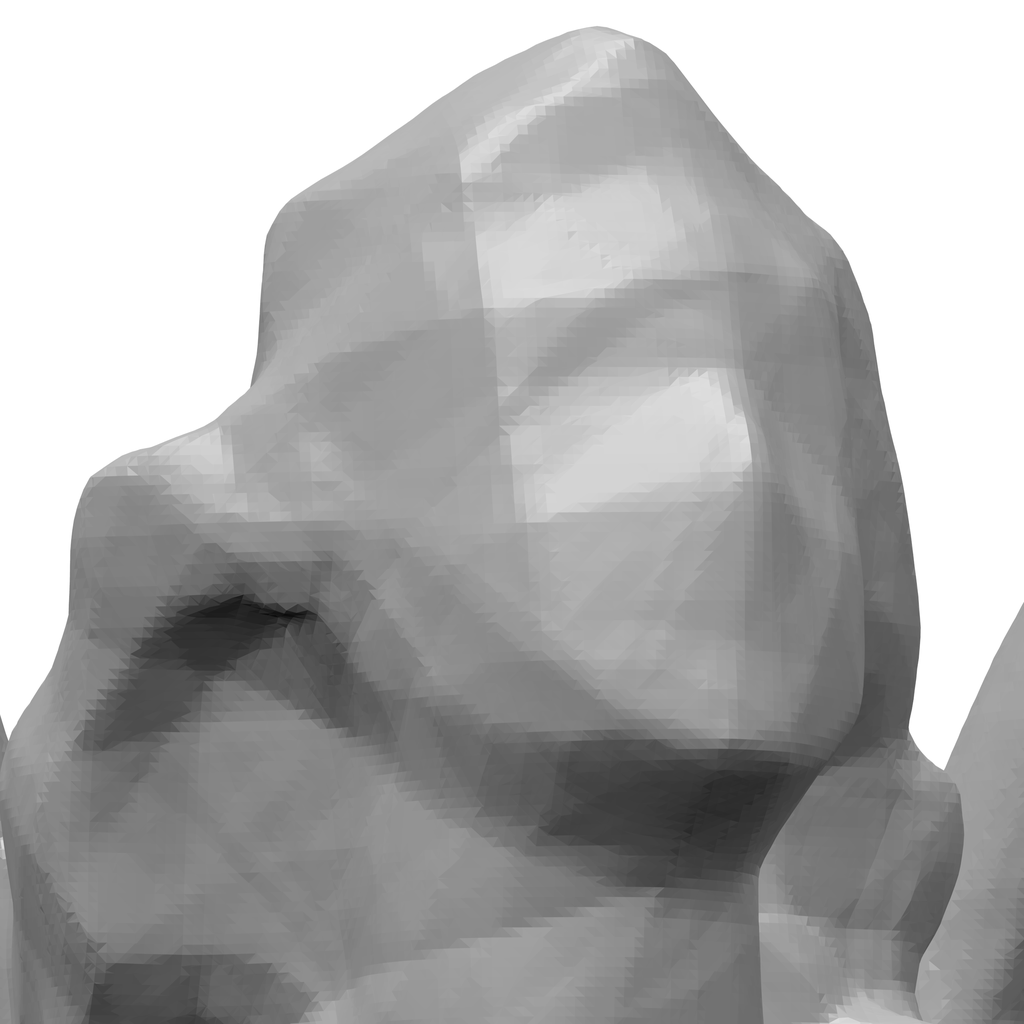}
            \mysubcaption{ReLU}{0.8969}{34 min}
        \end{subfigure}\hfill
        \begin{subfigure}{0.244\linewidth}
            \includegraphics[width=\linewidth]{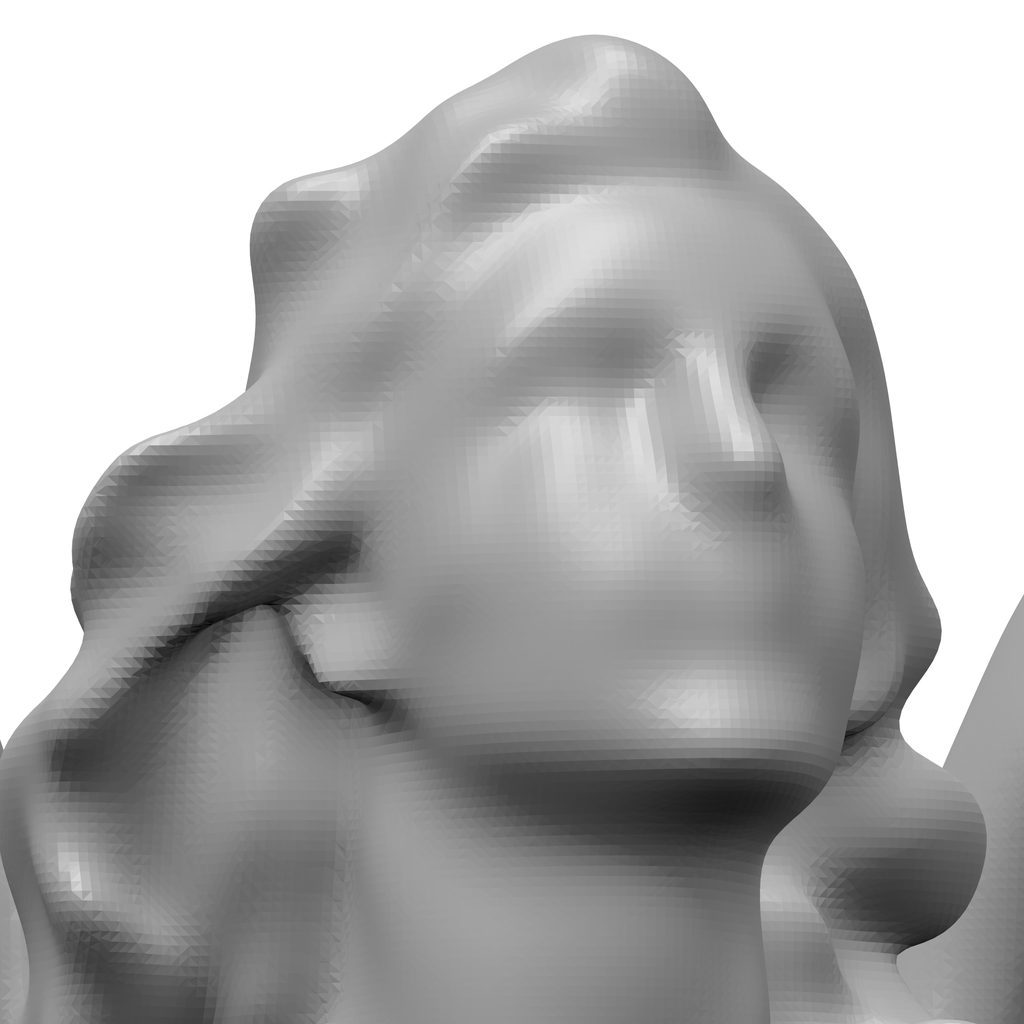}  
            \mysubcaption{SIREN}{0.7263}{34 min}      
        \end{subfigure}\hfill
        \begin{subfigure}{0.244\linewidth}
            \includegraphics[width=\linewidth]{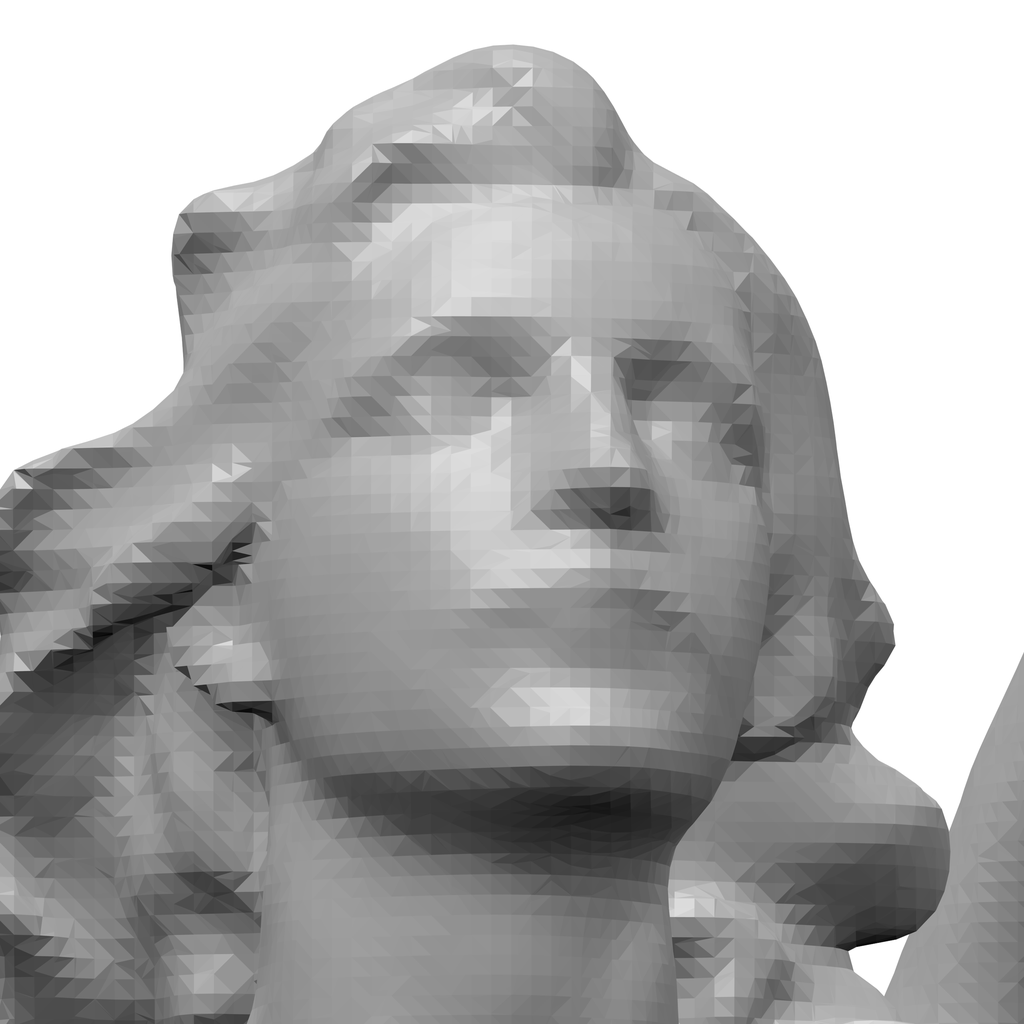}
            \mysubcaption{AdaHOSC}{0.9804}{35 min}        
        \end{subfigure}%
        
        \begin{subfigure}{0.244\linewidth}
            \includegraphics[width=\linewidth]{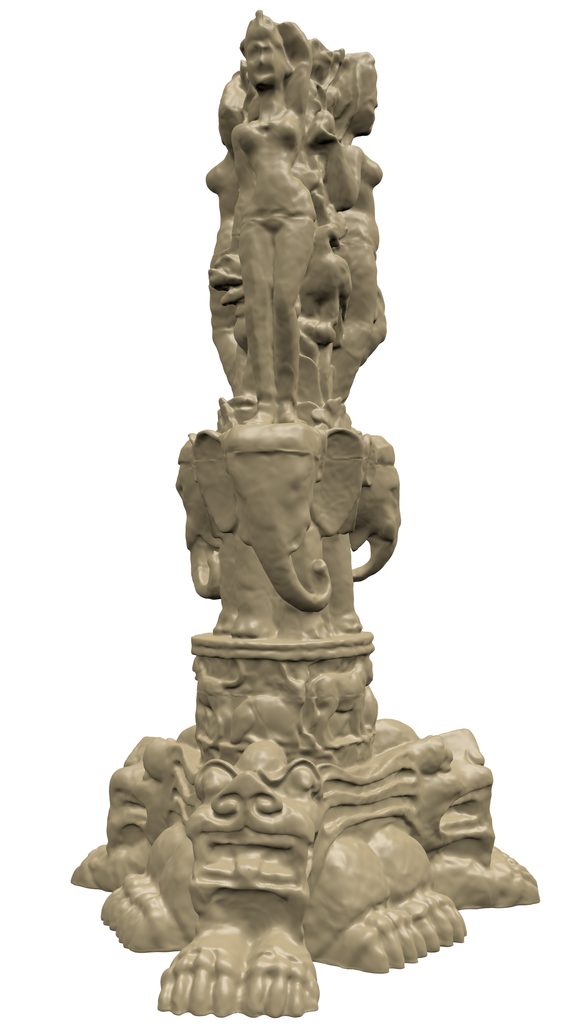}        
            \mysubcaption{Statuette}{IoU}{train time}        
        \end{subfigure}\hfill
        \begin{subfigure}{0.244\linewidth}
            \includegraphics[width=\linewidth]{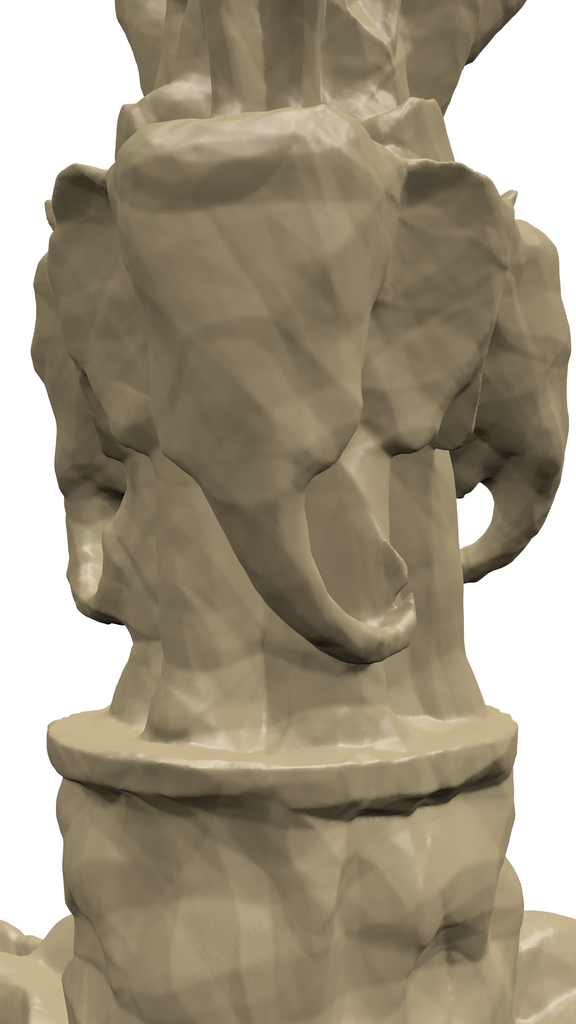}        
            \mysubcaption{ReLU}{0.8778}{34 min}
        \end{subfigure}\hfill
        \begin{subfigure}{0.244\linewidth}
            \includegraphics[width=\linewidth]{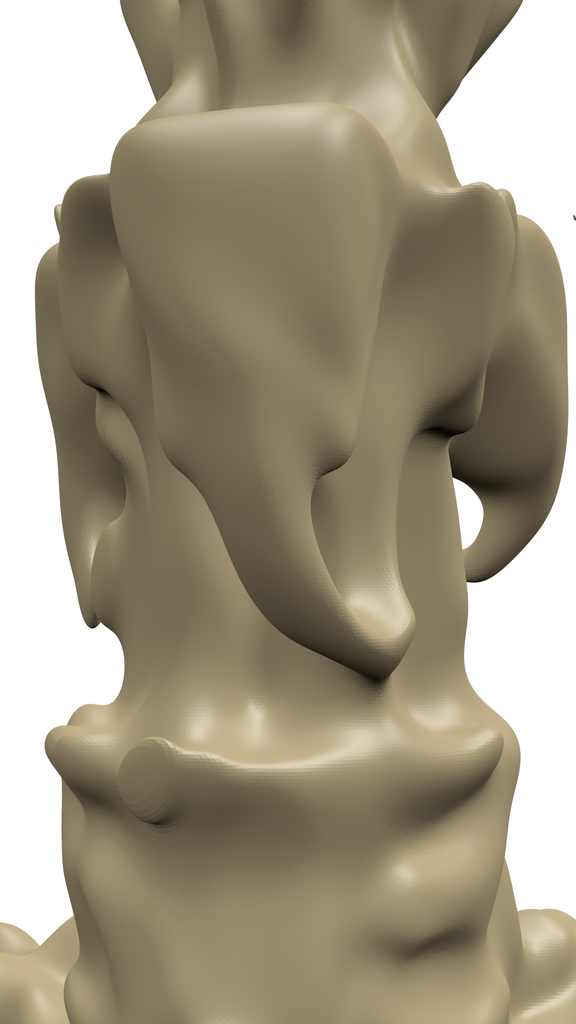}                
            \mysubcaption{SIREN}{0.8168}{34 min} 
        \end{subfigure}\hfill
        \begin{subfigure}{0.244\linewidth}
            \includegraphics[width=\linewidth]{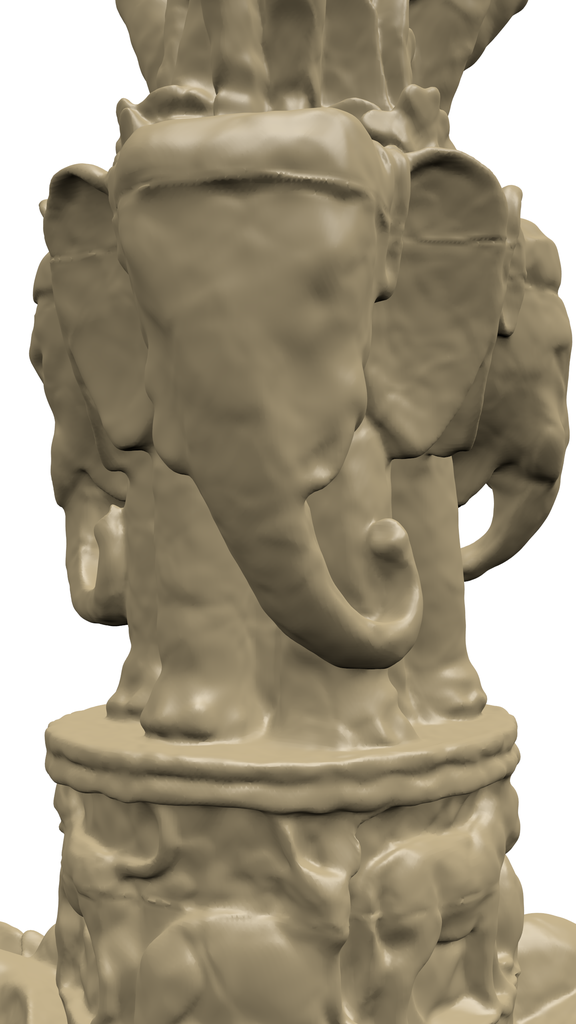}        
            \mysubcaption{AdaHOSC}{0.9587}{38 min} 
        \end{subfigure}%
    
        \begin{subfigure}{0.244\linewidth}
            \includegraphics[width=\linewidth,trim={296px 0 232px 0},clip]{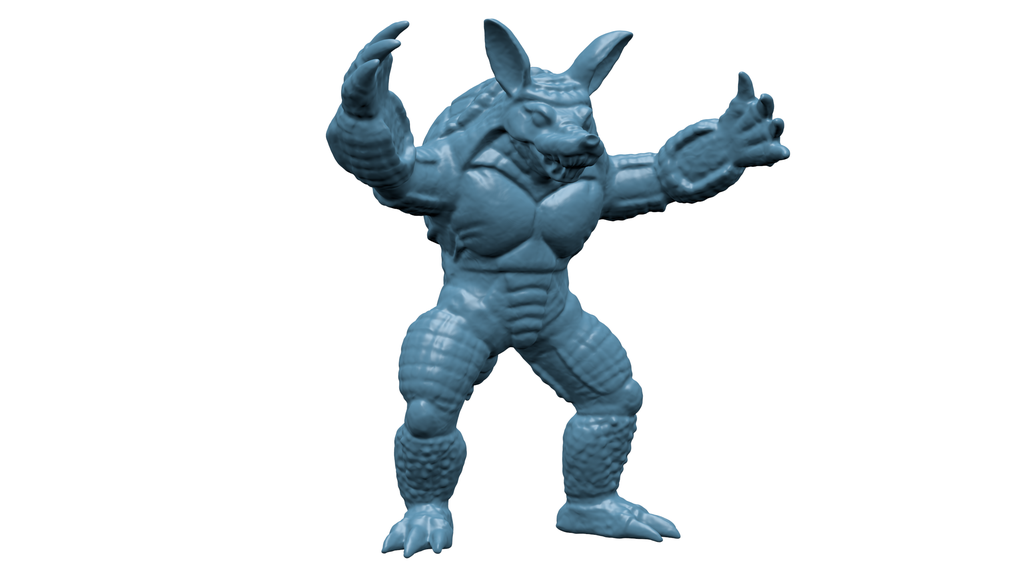}  
            \mysubcaption{Armadillo}{IoU}{train time}            
        \end{subfigure}\hfill
        \begin{subfigure}{0.244\linewidth}
            \includegraphics[width=\linewidth]{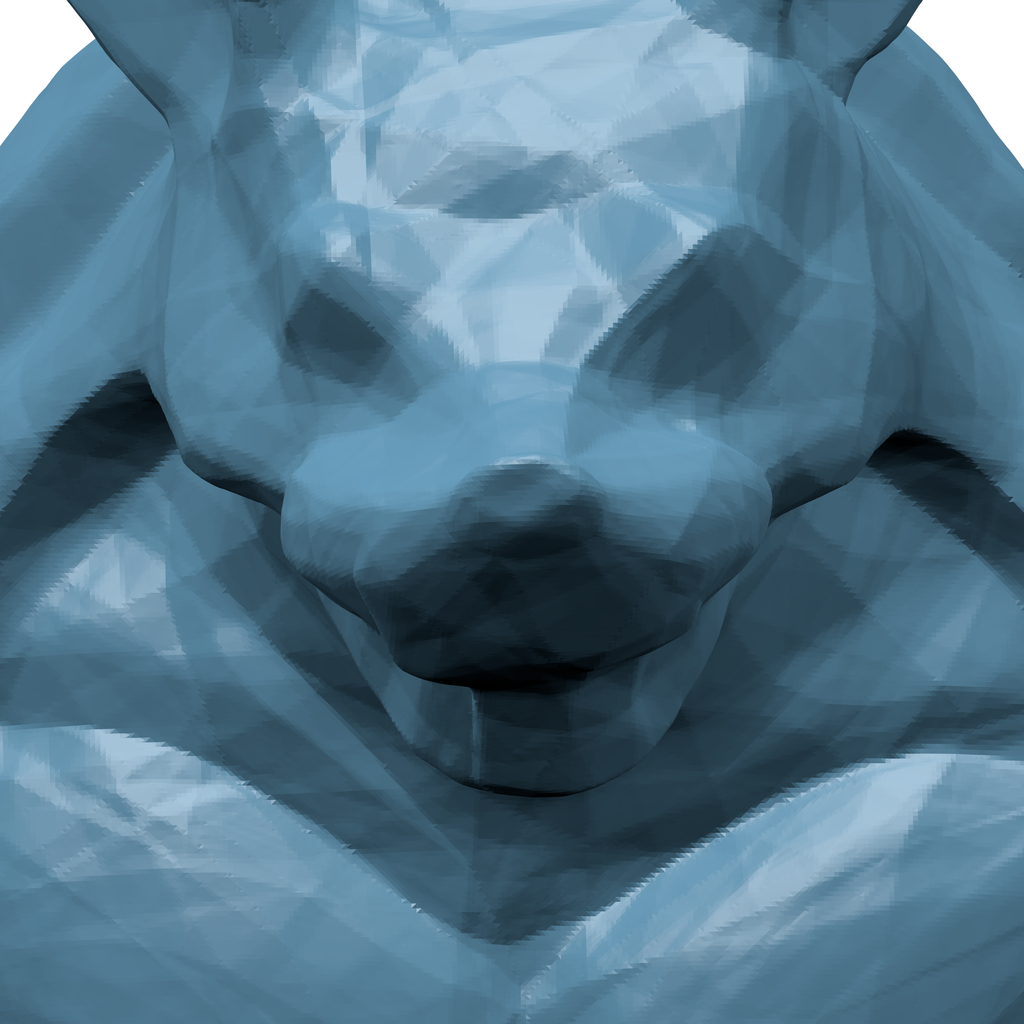}        
            \mysubcaption{ReLU}{0.9328}{35 min}
        \end{subfigure}\hfill
        \begin{subfigure}{0.244\linewidth}
            \includegraphics[width=\linewidth]{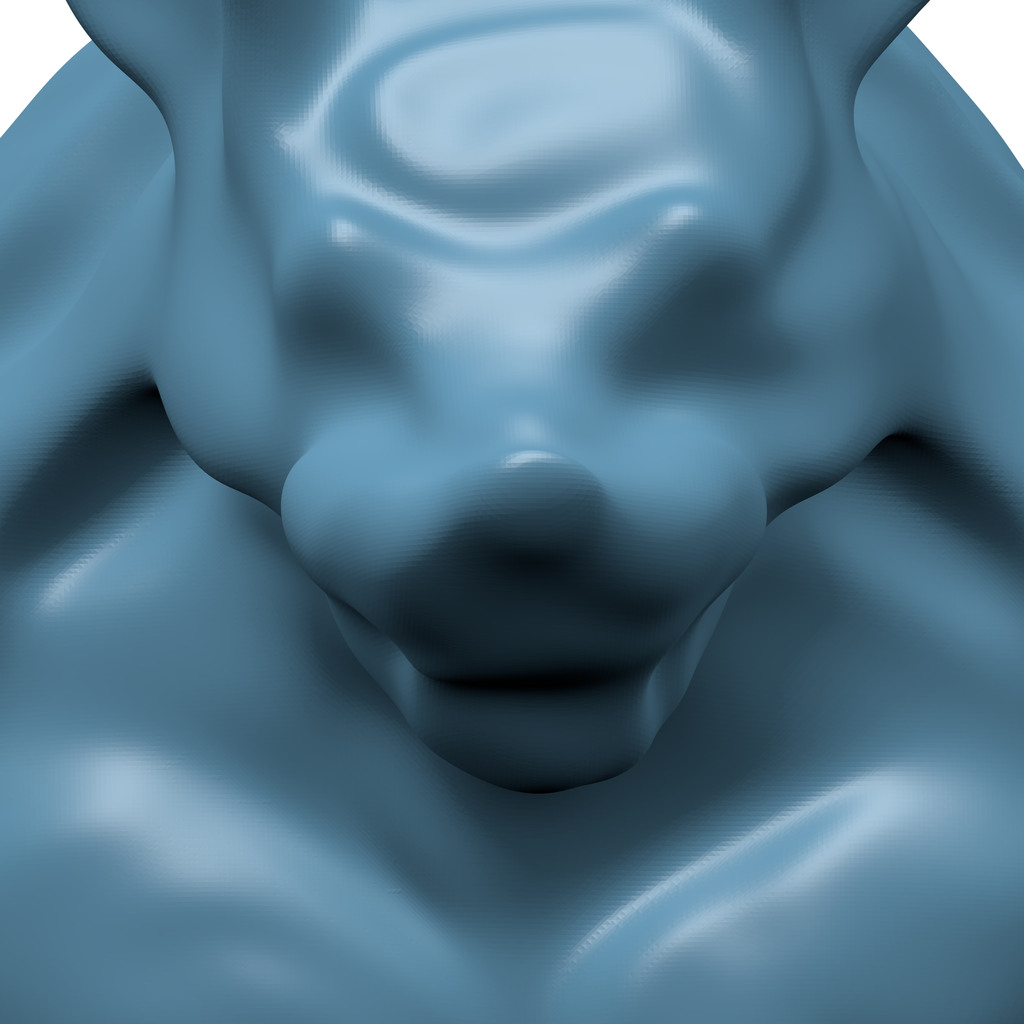}        
            \mysubcaption{SIREN}{0.9153}{36 min}
        \end{subfigure}\hfill
        \begin{subfigure}{0.244\linewidth}
            \includegraphics[width=\linewidth]{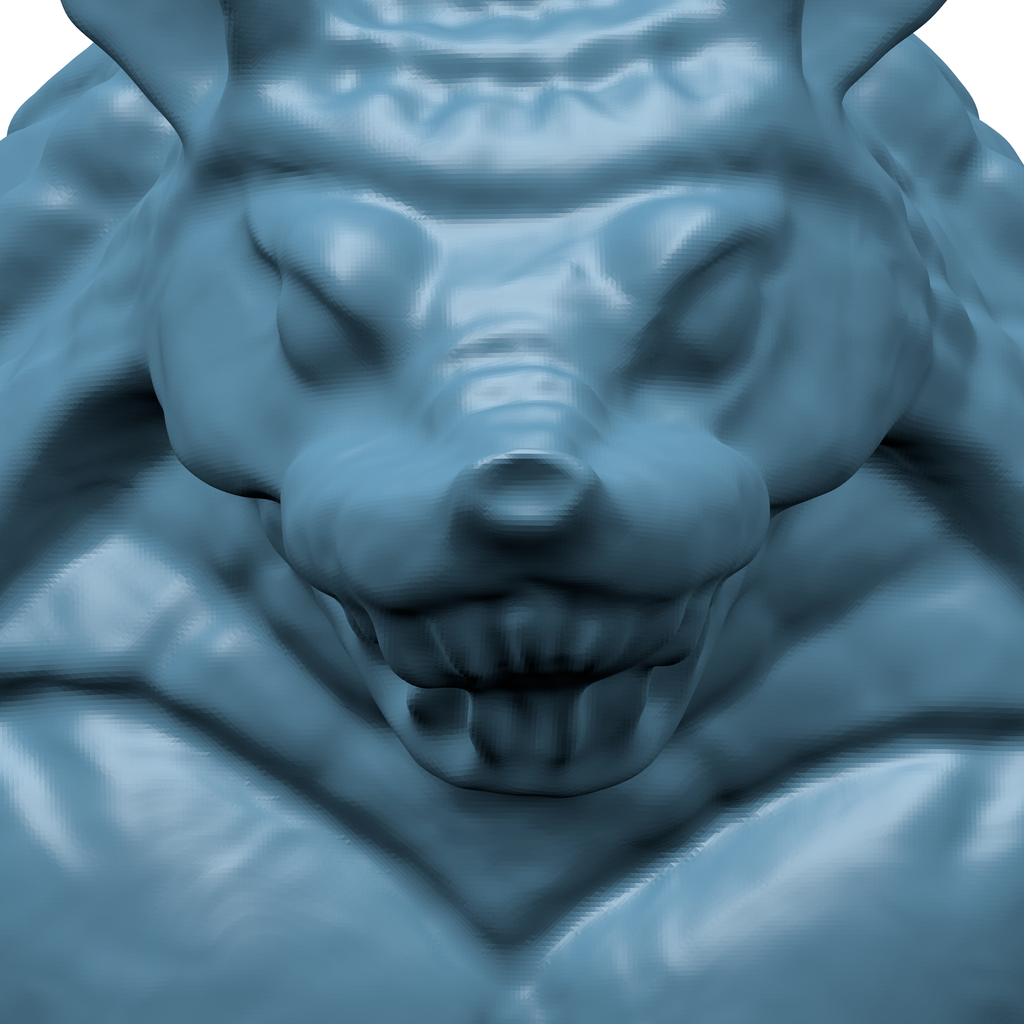}        
            \mysubcaption{AdaHOSC}{0.9775}{37 min}
        \end{subfigure}
    
        \begin{subfigure}{0.244\linewidth}
            \includegraphics[width=\linewidth]{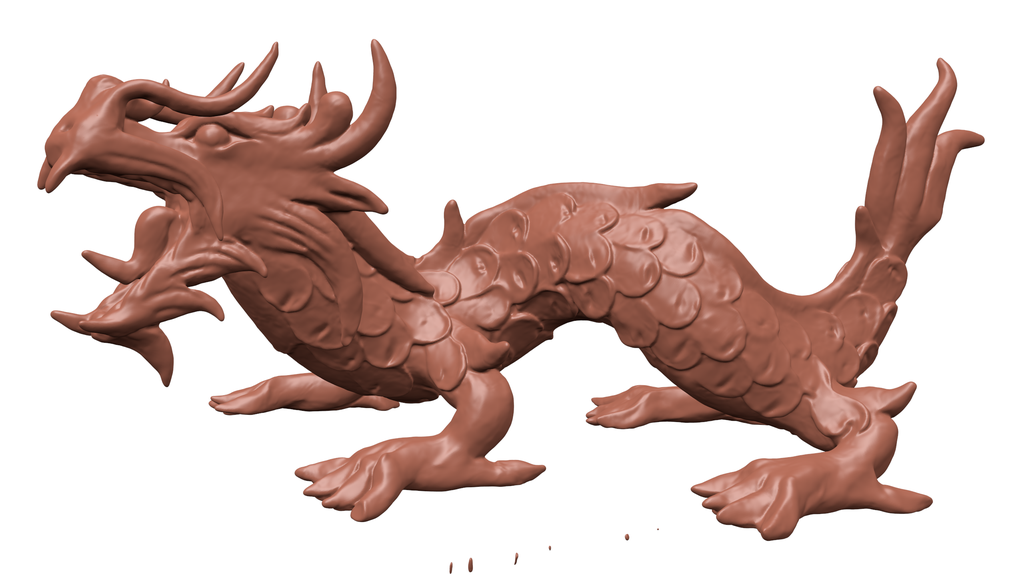}        
            \mysubcaption{Dragon}{IoU}{train time}        
        \end{subfigure}\hfill    
        \begin{subfigure}{0.244\linewidth}
            \includegraphics[width=\linewidth]{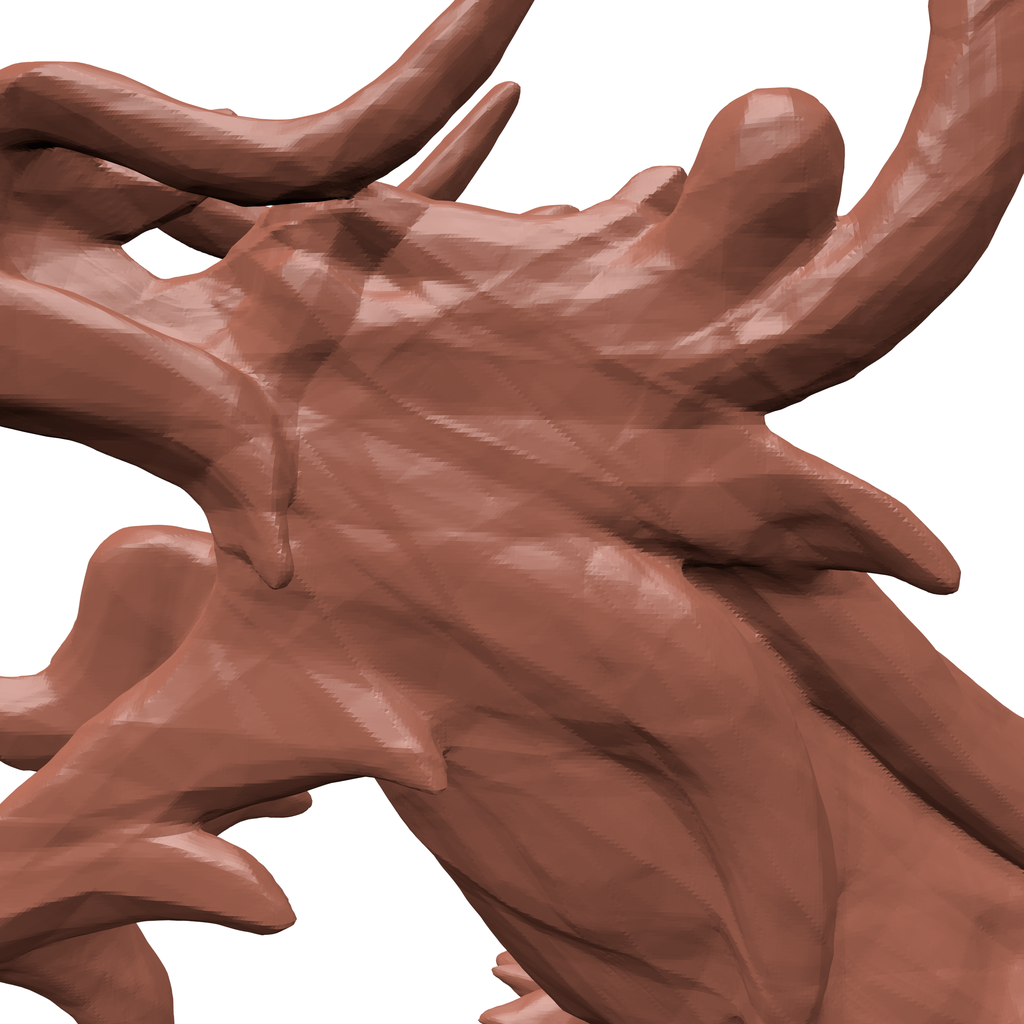}        
            \mysubcaption{ReLU}{0.8969}{34 min}
        \end{subfigure}\hfill
        \begin{subfigure}{0.244\linewidth}
            \includegraphics[width=\linewidth]{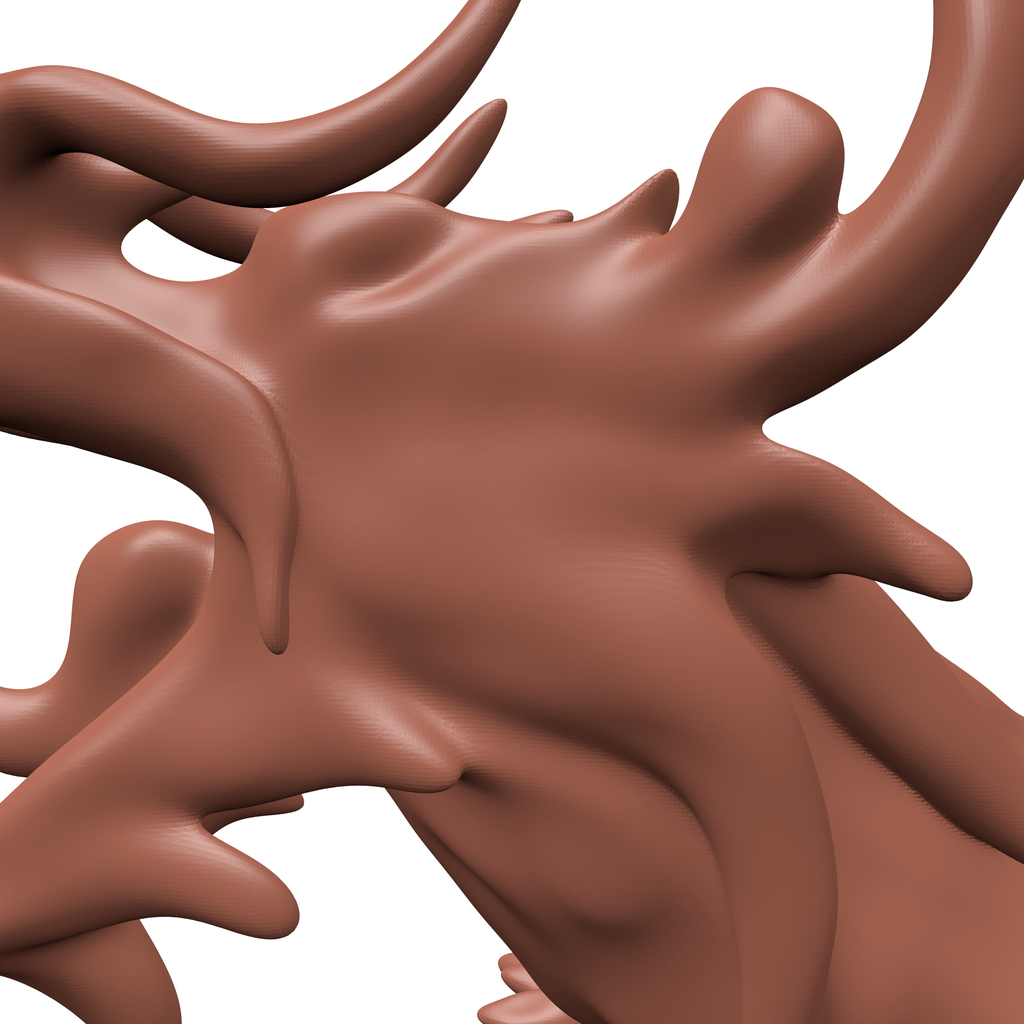}        
            \mysubcaption{SIREN}{0.8263}{34 min}
        \end{subfigure}\hfill
        \begin{subfigure}{0.244\linewidth}
            \includegraphics[width=\linewidth]{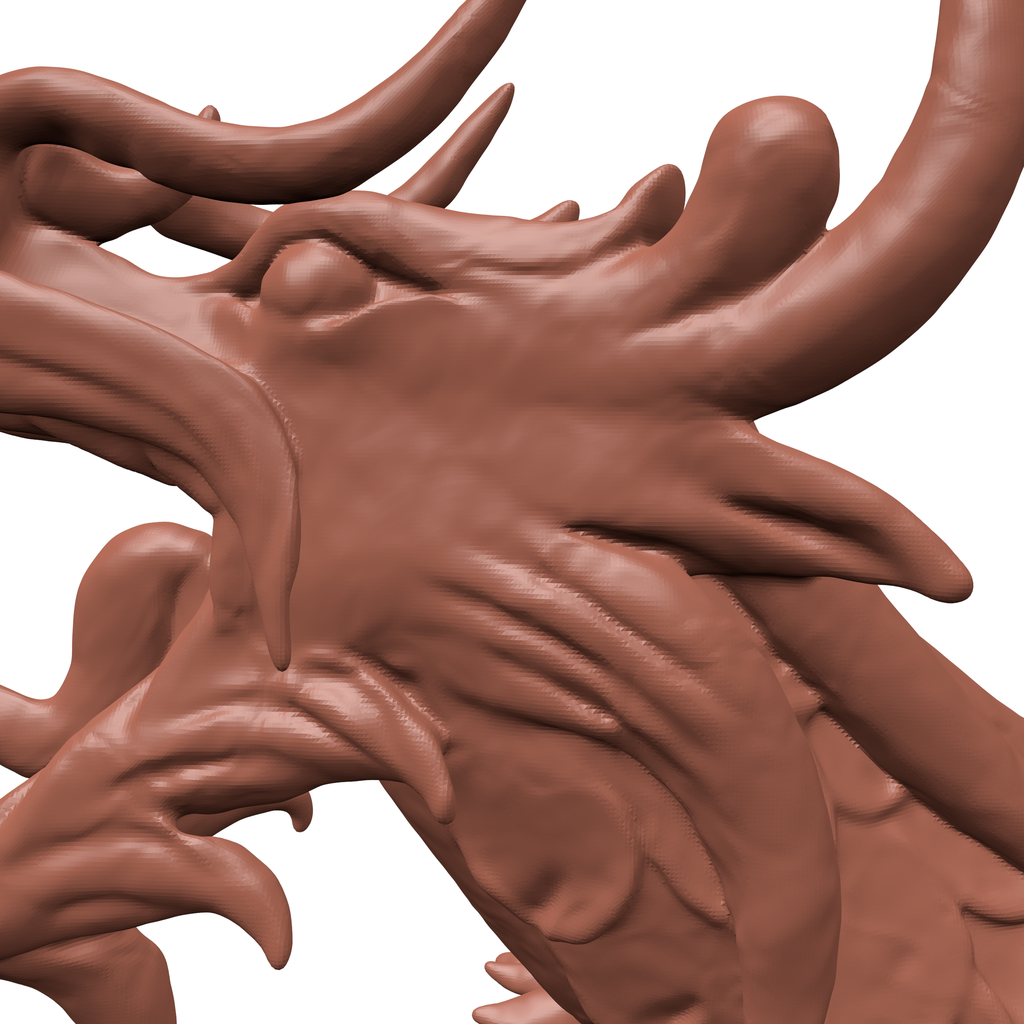}  
            \mysubcaption{AdaHOSC}{0.9628}{35 min}        
        \end{subfigure}
    
        \caption{
            Comparison of HOSC to other methods in 3D SDF reconstruction. All examples trained on 5-layer, $256$ hidden neurons model for 20 epochs. Note that HOSC is showing superior reconstruction quality, which is within the range of methods utilizing positional encoding, like Dictionary Fields~\citep{Chen2023FactorFA}.
            For the evaluation of the IoU, we used the dataset from \href{https://github.com/autonomousvision/factor-fields}{https://github.com/autonomousvision/factor-fields}. 
            Note that Lucy was evaluated only on a $512$ resolution mesh.
        }
        \label{fig:3d_sdf_1}
    \end{figure}

\bibliography{hosc_references}
\bibliographystyle{iclr2024_conference}

\end{document}